\documentclass[runningheads]{llncs}

\usepackage{eccv}

\usepackage{eccvabbrv}

\usepackage{graphicx}
\usepackage{subcaption}
\usepackage{amsmath,amsfonts}
\usepackage{bbding}
\usepackage{array}
\usepackage{longtable}
\usepackage{appendix}
\usepackage{textcomp}
\usepackage{stfloats}
\usepackage{verbatim}
\usepackage{color}
\usepackage{balance}
\usepackage[ruled,linesnumbered]{algorithm2e}
\usepackage{algpseudocode}
\usepackage{booktabs}
\usepackage{multirow}
\usepackage{hhline}
\usepackage{colortbl}
\usepackage{makecell}
\usepackage{float}

\usepackage{threeparttable} 
\usepackage{pifont} 

\UseRawInputEncoding
\usepackage[accsupp]{axessibility}  

\usepackage{hyperref}

\usepackage{orcidlink}

\begin{document}

\title{Preventing Catastrophic Overfitting in Fast Adversarial Training: A Bi-level Optimization Perspective} 

\titlerunning{Preventing Catastrophic Overfitting: A Bi-level Optimization Perspective}

\author{Zhaoxin~Wang\inst{1}\orcidlink{0009-0009-5860-8370} \and
Handing~Wang\inst{1}\textsuperscript{\Envelope}\orcidlink{0000-0002-4805-3780} \and
Cong~Tian\inst{2}\orcidlink{0000-0002-5429-4580} \and
Yaochu~Jin\inst{3}\orcidlink{0000-0003-1100-0631}}

\authorrunning{Z.~Wang et al.}

\institute{School of Artificial Intelligence, Xidian University, Xi'an, China \\ \email{zxwang74@163.com} \and School of Computer Science and Technology, Xidian University, Xi'an, China \\ \email{ctian@mail.xidian.edu.cn} \and School of Engineering, Westlake University, Zhejiang Hangzhou, China \\ \email{jinyaochu@westlake.edu.cn}}

\maketitle

\let\thefootnote\relax\footnotetext{\textsuperscript{\Envelope} Corresponding Author: hdwang@xidian.edu.cn}

\begin{abstract}
 Adversarial training (AT) has become an effective defense method against adversarial examples (AEs) and it is typically framed as a bi-level optimization problem. Among various AT methods, fast AT (FAT), which employs a single-step attack strategy to guide the training process, can achieve good robustness against adversarial attacks at a low cost. However, FAT methods suffer from the catastrophic overfitting problem, especially on complex tasks or with large-parameter models. In this work, we propose a FAT method termed FGSM-PCO, which mitigates catastrophic overfitting by averting the collapse of the inner optimization problem in the bi-level optimization process. FGSM-PCO generates current-stage AEs from the historical AEs and incorporates them into the training process using an adaptive mechanism. This mechanism determines an appropriate fusion ratio according to the performance of the AEs on the training model. Coupled with a loss function tailored to the training framework, FGSM-PCO can alleviate catastrophic overfitting and help the recovery of an overfitted model to effective training. We evaluate our algorithm across three models and three datasets to validate its effectiveness. Comparative empirical studies against other FAT algorithms demonstrate that our proposed method effectively addresses unresolved overfitting issues in existing algorithms.
 
\end{abstract}

\section{Introduction}

The security of deep neural networks \cite{chakraborty2018adversarial, li2022backdoor, liu2020privacy} has raised increasing concerns as they always face malicious threats \cite{szegedy2013intriguing, goodfellow2014explaining, gu2017badnets, shokri2017membership, zhu2019deep}. Among these threats, adversarial examples (AEs) pose a significant risk by causing neural networks to make an incorrect prediction or classification without participating in the training phase \cite{moosavi2016deepfool, chen2017zoo, brendel2017decision, ilyas2018black, duan2021adversarial}. Adversarial training (AT) \cite{madry2017towards, wang2023adversarial, wang2023better, wang2019improving, DinghuaiZhang2019YouOP, shafahi2019adversarial} is one of the most effective ways to resist this adversarial attack, which can significantly enhance the models' classification ability for AEs rather than merely identifying them. AT can be framed as a bi-level optimization problem \cite{madry2017towards} as follows:
\begin{equation}
    \label{AT}
    \min _{\theta} \underset{(\boldsymbol{x}, \boldsymbol{y}) \sim \mathcal{D}}{\mathbb{E}}\left[\max _{\left|\boldsymbol{\delta}\right|_p \leq \epsilon} \mathcal{L}\left(\boldsymbol{f}_{\theta}\left(\boldsymbol{x+\delta}\right), \boldsymbol{y}\right)\right],
\end{equation}
where $\boldsymbol{f}_{\theta}$ represents a deep model with parameters $\mathbf{\theta}$, $\mathcal{D}$ represents the data distribution of clean examples $\boldsymbol{x}$ and their labels $\boldsymbol{y}$, $\boldsymbol{\delta}$ denotes the perturbation, $\epsilon$ denotes the perturbation threshold, and $|\cdot|_p$ is defined as $l_p$ norm.

Typically, the univariate search technique is commonly employed for this bi-level optimization problem \cite{colson2007overview}, where attack strategies generate AEs with fixed model parameters in the inner maximum process, and these AEs are used to update the model parameters in the outer minimum optimization process. However, this maximum process requires multiple iterative processes to find an effective $\boldsymbol{\delta}$, leading to significant computation demands.

\begin{figure}[htbp]
	\begin{subfigure}[b]{0.48\textwidth}
		\includegraphics[width=\linewidth]{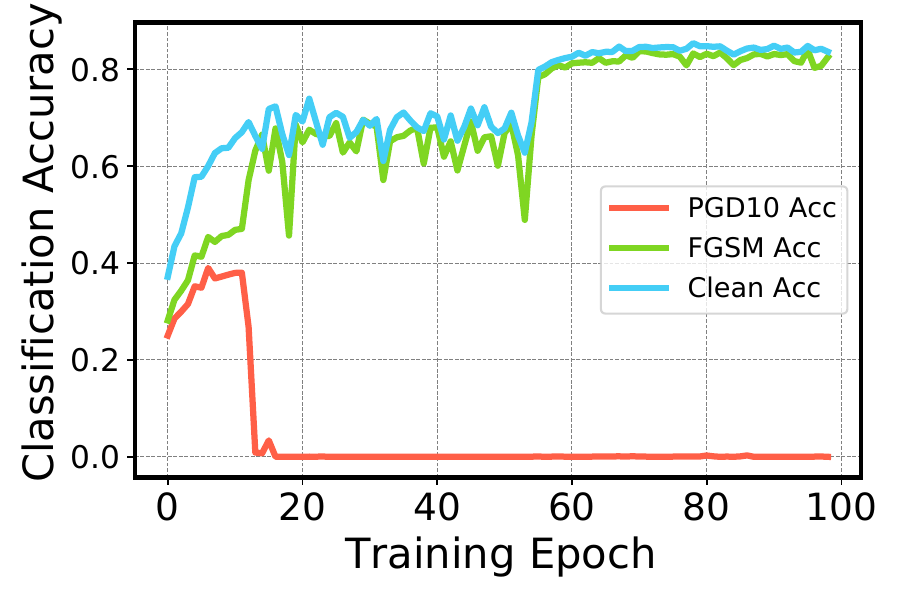}
		\caption{CIFAR10 on ResNet18}
		\label{CO_cifar}
	\end{subfigure}
	\hfill
	\begin{subfigure}[b]{0.48\textwidth}
		\includegraphics[width=\linewidth]{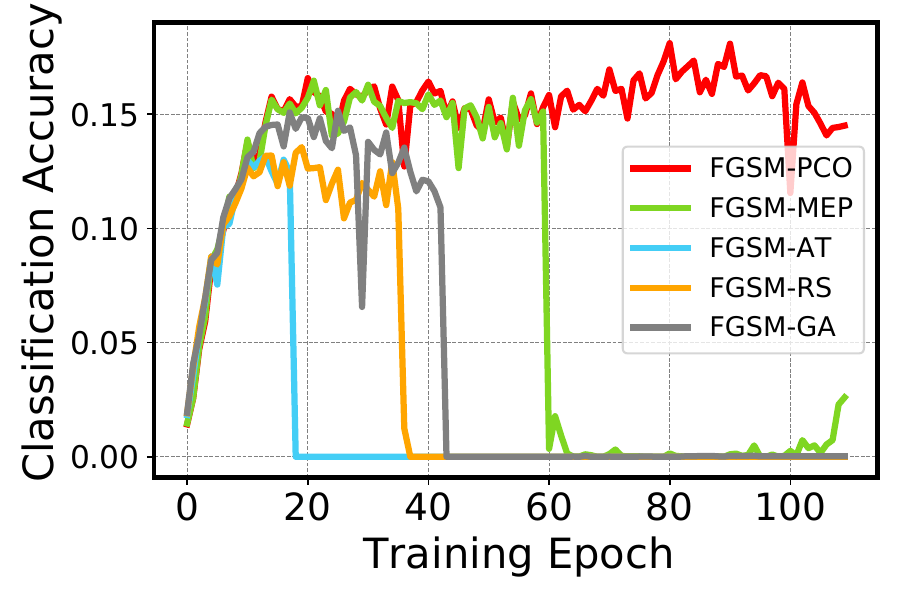}
		\caption{Tiny-ImageNet on PreActResNet18}
		\label{CO}
	\end{subfigure}
	\caption{Catastrophic overfitting phenomenon in FAT. (a) is on the CIFAR10 dataset with a multi-step learning rate. (b) shows the overfitting on the Tiny-ImageNet dataset with a cyclic learning rate. Most FAT algorithms cannot prevent catastrophic overfitting.}
	\label{fig:figures}
\end{figure}

To accelerate the training process of AT, fast adversarial training (FAT) methods have been developed \cite{goodfellow2014explaining, EricWong2020FastIB, MaksymAndriushchenko2020UnderstandingAI, YihuaZhang2022RevisitingAA, XiaojunJia2022PriorGuidedAI, kim2021understanding}, which utilize the fast gradient sign attack method (FGSM) \cite{goodfellow2014explaining}, a single-step attack method, as the inner attack strategy. However, coupling the univariate search technique with FGSM is highly susceptible to the collapse of FAT, leading to the catastrophic overfitting problem where the classification accuracy of AEs suddenly drops to 0\% under multi-step attacks like Projected Gradient Descent (PGD) attack. Fig.~\ref{CO_cifar} shows an example of catastrophic overfitting observed in FGSM-AT \cite{goodfellow2014explaining}.

To address the catastrophic overfitting problem, FGSM-RS \cite{EricWong2020FastIB} initializes the perturbation with a random uniform distribution $U[-\epsilon,\epsilon]$ and adopts a large attack step size in FGSM strategy. Zhang \textit{et al.} \cite{MaksymAndriushchenko2020UnderstandingAI, YihuaZhang2022RevisitingAA} propose integrating regularization into the loss function to mitigate the issue. FGSM-MEP \cite{XiaojunJia2022PriorGuidedAI} proposes that the previous perturbation can guide the perturbation initialization in FAT, leveraging the accumulation of the perturbation momentum to overcome the catastrophic overfitting problem. These methods significantly improve the performance of FAT, allowing FAT methods to achieve comparable robustness to the multi-step attack-guided AT at a low cost. 

Nonetheless, these methods are not immune to catastrophic overfitting on more complex tasks \cite{deng2009imagenet} or with large-parameter models. Fig.~\ref{CO} shows catastrophic overfitting phenomenon under different FAT methods. Compared with FGSM-AT, these algorithms primarily delay the onset of overfitting but cannot completely prevent it. The reason for this catastrophic overfitting phenomenon is the collapse of the inner optimization problem, and the collapse derives from coupling FGSM this single, large-step attack strategy with the alternate optimization method between the inner and outer problem. As shown in Fig.~\ref{CO_cifar}, the classification accuracy under FGSM attack steep rise. Once this inner optimization fails, the entire optimization process becomes ineffective. Besides, existing FAT methods lack mechanisms to rectify this flaw. 

Therefore, we propose a FAT algorithm termed FGSM-PCO, which can effectively address the catastrophic overfitting problem in FAT. Specifically, we fuse the historical and current perturbations for training according to an adaptive mechanism. This adaptive mechanism can adjust the ratio of the two types of AEs according to the performance on the model. It ensures that FGSM-PCO can correct the training course, avoiding the collapse of the inner optimization problem. In addition, we propose a regularization to assist this methodology, which advocates a consistent prediction for training samples before and after fusion. In summary, our main contributions are:

\begin{itemize}
    \item  We propose a FAT framework that incorporates both the historical and current perturbation into training through an adaptive mechanism. This framework significantly mitigates the issue of catastrophic overfitting and helps the recovery of an overfitted model to effective training.
    \item  A tailored regularization strategy is proposed to prevent the collapse of the inner optimization problem within the FAT framework, which advocates consistent predictions for samples both before and after the fusion.
    \item  Experimental results show that our algorithm improves the classification accuracy on both clean examples and AEs.
\end{itemize}

\section{Related work}
\subsection{Fast Adversarial Training}
Most AT methods \cite{ChongliQin2019AdversarialRT, SeyedMohsenMoosaviDezfooli2019RobustnessVC, ChengzhiMao2019MetricLF, wang2023adversarial} can be formulated as a bi-level optimization problem as shown in Eq.(\ref{AT}). In the AT framework, a multi-step attack method PGD is frequently employed to generate a high-quality $\delta$ in the inner optimization process. PGD can be described as:
\begin{equation}
	\label{PGD}
	\boldsymbol{x}_{t}^*=\boldsymbol{x}+\Pi_{[-\epsilon, \epsilon]}\left[\boldsymbol{\delta}_{t-1} + \alpha \operatorname{sign}\left(\nabla_{\boldsymbol{x}} \mathcal{L}(\boldsymbol{f}_{\theta}(\boldsymbol{x}_{t-1}^*), \boldsymbol{y})\right)\right],
\end{equation}
where $\Pi_{[-\epsilon, \epsilon]}$ represents the projection that limits the input to the range $[-\epsilon, \epsilon]$, $\boldsymbol{\delta}_0$ is the randomly initialized perturbation, $\boldsymbol{x}$ are the clean examples, $\alpha$ is the attack step size.

This multi-step attack strategy makes AT requires high computational cost, which is impractical in some tasks. Therefore, researchers propose FAT, which employs FGSM, a single-step attack method instead of the PGD method, to generate the training examples. FGSM can be described as:
\begin{equation}
	\label{FGSM}
	\boldsymbol{x}^*=\boldsymbol{x}+\epsilon  \operatorname{sign}\left(\nabla_{\boldsymbol{x}} \mathcal{L}(\boldsymbol{f}_{\theta}(\boldsymbol{x+\delta}), \boldsymbol{y})\right),
\end{equation}
where $\boldsymbol{\delta}$ is the perturbation initialization. Compared with the PGD attack, FGSM applies a single large step, using the perturbation threshold $\epsilon$ directly as the step size, which significantly accelerates the training process. Despite its efficiency, FGSM increases the risk of FAT encountering the collapse of the inner optimization problem, leading to catastrophic overfitting.

\subsection{Dilemma in Bi-Level Optimization Problems}
\begin{figure}[htbp]
	\centering
	\includegraphics[width=1\linewidth]{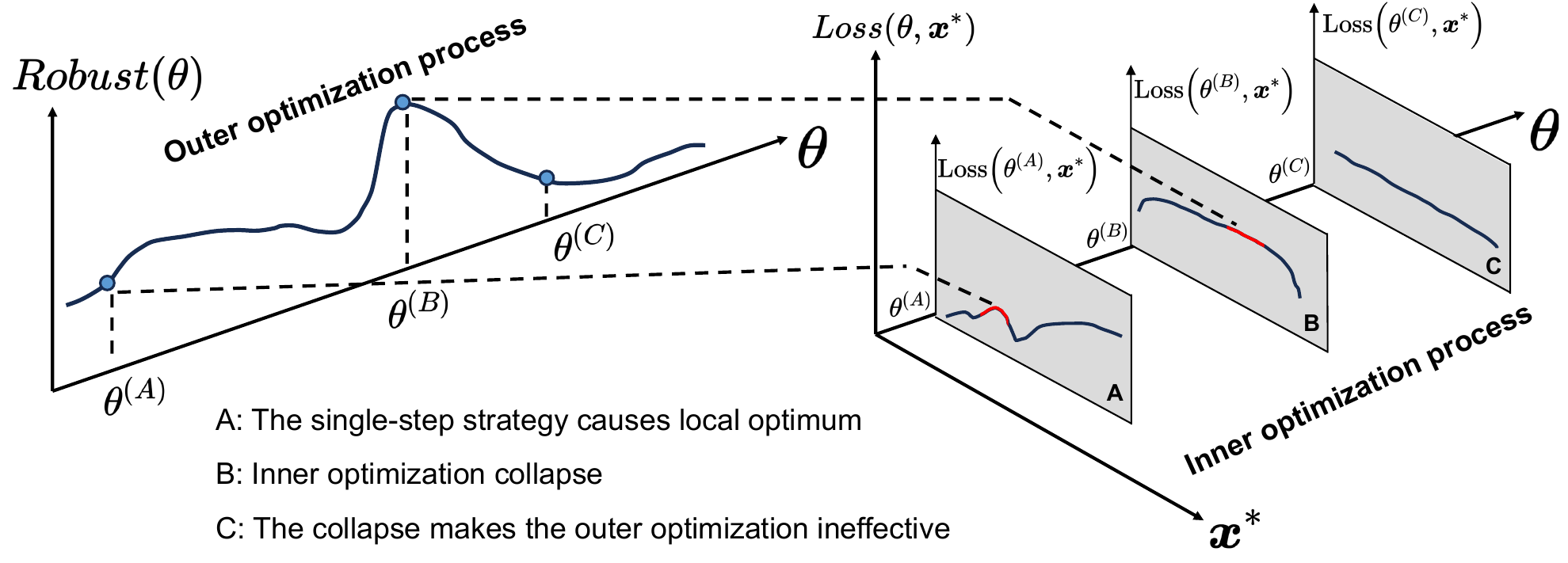}
	\caption{Optimizing the inner and outer problems alternately is easy to cause the collapse of the bi-level optimization.}
	\label{bilevel}
\end{figure}

Bi-level optimization involves two interconnected optimization problems, where the solution to the outer-level problem depends on the outcome of the inner-level problem \cite{dempe2015bilevel} and vice versa. This nested structure can lead to conflicts and potential collapse, particularly when the inner and outer levels are optimized alternately. In the FAT framework, the bi-level optimization process can be described as Fig.~\ref{bilevel}.

In the bi-level optimization framework, the objective function of the outer layer $Robust(\theta)$, solely depends on the outer variable $\theta$, while the inner-level objective function, $Loss(\theta,\boldsymbol{x}^*)$, is dependent on both the outer variable $\theta$ and the inner variable $\boldsymbol{x}^*$. Given an outer variable $\theta^{(A)}$, we obtain the corresponding loss surface on the right side in Fig.~\ref{bilevel}. Once the inner optimization collapse occurs under $\theta^{(B)}$, the nested structure of the bi-level optimization exacerbates the collapse phenomenon in optimization. In the FAT, alternating optimization of inner and outer levels, coupled with FGSM's large step size and constrained perturbation search space, increases the risk of collapse in bi-level optimization.

\subsection{Prior-Guided Fast Adversarial Training}
FGSM-MEP \cite{XiaojunJia2022PriorGuidedAI} proposes that prior knowledge of adversarial perturbations can effectively guide subsequent AT. This method employs the accumulation of the gradient momentum over all previous training epochs to initialize the FGSM perturbation $\boldsymbol{\eta}$. The process is detailed as follows:
\begin{equation}
	\label{MEP}
	\begin{aligned}
		& \mathbf{g}_c=\operatorname{sign}\left(\nabla_{\mathbf{x}} \mathcal{L}\left(\boldsymbol{f}_{\theta}\left(\boldsymbol{x}+\boldsymbol{\eta}_{E_t} \right), \mathbf{y}\right)\right),  \\
		& \mathbf{g}_{E_{t+1}}=\mu \cdot \mathbf{g}_{E_t}+\mathbf{g}_c, \\
		& \boldsymbol{\delta}_{E_{t+1}}=\Pi_{[-\epsilon, \epsilon]}\left[\boldsymbol{\eta}_{E_t}+\alpha \cdot \mathbf{g}_c\right], \\
		& \boldsymbol{\eta}_{E_{t+1}}=\Pi_{[-\epsilon, \epsilon]}\left[\boldsymbol{\eta}_{E_t}+\alpha \cdot \operatorname{sign}\left(\mathbf{g}_{E_{t+1}}\right)\right],
	\end{aligned}
\end{equation}
where $\mathbf{g}_{E_{t+1}}$ represents the gradient momentum in the $(t+1)$-th epoch, $\boldsymbol{\delta}$ is the perturbation and $\mu$ is the decay factor. 

While this approach significantly enhances the performance of FAT, it still has some issues. Although the accumulation of momentum provides prior knowledge for AT, as training progresses, continuous accumulation of gradient momentum can inadvertently increase the divergence between adversarial and clean examples, potentially degrading the model's performance on clean data, even though the momentum is reset periodically during training. Furthermore, while leveraging momentum to initialize subsequent perturbations decreases the risk of ineffective optimization and mitigates catastrophic overfitting, it fails to rectify the algorithm once the inner-level optimization collapses. In addition, although the accumulation of momentum does not increase the algorithm's runtime, it significantly raises memory requirements.

\section{Proposed Method}
To prevent the collapse of the inner optimization problem in FAT, we propose the FAT method FGSM-PCO. It generates AEs based on those from the previous training epoch and then fuses these two types of AEs based on their performance on the model. The fusion AEs are then utilized for training within a framework guided by a novel loss function, ensuring FGSM-PCO can rectify the trend toward inner-layer optimization collapse. Specifically, when AEs from the current stage prove ineffective for FAT, the adaptive fusion mechanism prioritizes the prior-stage AEs, thus maintaining the training efficacy.

\subsection{Procedure of Proposed Method}
As illustrated in Fig.~\ref{CO_cifar}, catastrophic overfitting is marked by a sudden spike in the model's accuracy on FGSM-generated AEs, signaling the collapse of the inner optimization within FAT. This collapse makes the AEs generated by FGSM ineffective for current training. To prevent this and correct the trend when it emerges, the fusion process is skewed towards retaining a higher proportion of previous-stage AEs, thereby averting the entrapment of bi-level optimization in local minima and the subsequent collapse of inner optimization. The training process can be specifically described by Eq.(\ref{Procedure}).
\begin{equation}
    \label{Procedure}
    \begin{aligned}
        &\boldsymbol{x}_{\text{train}} = \lambda_t \boldsymbol{x}_{t-1}^* + (1 - \lambda_t) \boldsymbol{x}_{\text{am}}^*, \\
        &\boldsymbol{x}_{\text{am}}^* = \boldsymbol{x}_{t-1}^* + \gamma \boldsymbol{\delta}_t, \\
        &\boldsymbol{\delta}_t = \epsilon \mathbf{g}_t, \\
        &\mathbf{g}_t = \operatorname{sign}\left(\nabla_{\mathbf{x}} \mathcal{L}\left(\boldsymbol{f}_{\theta}\left(\boldsymbol{x}_{t-1}^*\right), \mathbf{y}\right)\right),\\
        &\boldsymbol{x}^*_0 = \boldsymbol{x} + U[-\epsilon, \epsilon], 
    \end{aligned}
\end{equation}
where $\boldsymbol{x}$ and $\boldsymbol{y}$ represent the clean examples and their corresponding labels, $\boldsymbol{x}_{t-1}^*$ are the AEs in the $(t-1)-th$ epoch, $\boldsymbol{x}_{am}^*$ are the current stage AEs with a amplified delta $\gamma \boldsymbol{\delta}_t$, $\gamma$ is the amplification factor, $\lambda_t$ represents the fusion factor, $U[-\epsilon, \epsilon]$ represents a uniform distribution. The detailed description is presented in Algorithm \ref{algorithm}.

\subsection{Adaptive Fusion Ratio}
Different from other FAT methods, which utilize FGSM to generate perturbations and attach them to the clean examples, we scale the perturbations with an amplification factor $\gamma$ as mentioned in Eq.(\ref{Procedure}). This amplification factor ensures the effectiveness of perturbations after fusion, $i.e.$, it balances the perturbation attenuation brought by the fusion faction $\lambda_t$.

To determine a good fusion factor $\lambda_t$, which can make the algorithm escape the collapse of the inner optimization problem. The fusion factor value is based on the performance of the AEs on the model. Specifically, through the model's classification confidence for the ground-truth category, which directly influences $\lambda_t$. The formula for $\lambda_t$ is:

\begin{equation}
	\label{adaptive ratio}
	\begin{gathered}
		\lambda_t = \boldsymbol{f}^i_{\theta}(\boldsymbol{x_{t}^*}) \\[1mm]
	\end{gathered}\\
\end{equation}
where $\boldsymbol{f}^i_{\theta}(\boldsymbol{x}_{t}^*)$ is the classification confidence, $i$ is the index of the ground-truth category. As shown in Eq.(\ref{Procedure}), when the catastrophic overfitting occurs, the training examples $\boldsymbol{x}_{train}$ skewed towards retaining a higher proportion of previous-stage AEs $\boldsymbol{x}_{t-1}^*$. This adaptive fusion mechanism ensures that FGSM-PCO can correct the catastrophic overfitting trend before the collapse of the inner optimization problem.

\subsection{Regularization Loss}
To better guide our proposed optimization process, we introduce a new loss function. For fusion AEs generated by Eq.(\ref{adaptive ratio}), we hope to achieve prediction results consistent with both the previous and the current AEs. Thus, apart from utilizing cross-entropy loss to direct AT, we propose a regularization strategy as defined in Eq.(\ref{regularization}).

\begin{equation}
	\label{regularization}
	\begin{gathered}
		\mathcal{L}_{PCO}(\boldsymbol{x}_{train},\boldsymbol{y}) = \mathcal{L}_{CE}(\boldsymbol{f}_{\theta}(\boldsymbol{x}_{train}),\boldsymbol{y}) \\
		+ \beta \big[ \mathcal{L}_1(\boldsymbol{f}_{\theta}(\boldsymbol{x}_{t}^*),\boldsymbol{f}_{\theta}(\boldsymbol{x}_{t-1}^*)) - \mathcal{L}_1(\boldsymbol{f}_{\theta}(\boldsymbol{x}_{train}),\boldsymbol{f}_{\theta}(\boldsymbol{x}_{t}^*)) \big],
	\end{gathered}
\end{equation}
where $\beta$ is the trade-off parameter, the first item represents the cross-entropy loss on the fusion AEs and the second item $\mathcal{L}_1$ represents the mean square loss with $l_2$ norm.

\begin{algorithm}
	\caption{FGSM-PCO}
	\label{algorithm}
	\KwIn{Target model $\boldsymbol{f}_{\theta}$, maximum perturbation $\epsilon$, training epoch $T$, mini-batch data $\mathbb{B}$, learning rate $\alpha$, amplification parameter $\gamma$.}
	
	\KwOut{Model weights ${\theta}$}
	
	\For{t = 1,...,T}{
		\For{$\{\boldsymbol{x},\boldsymbol{y}$\}$\sim \mathbb{B}$}{
			\If{t == 1}
			{
				$\boldsymbol{x}_{t-1}^* \gets \boldsymbol{x} + U[-\epsilon,\epsilon]$;
			}
		
		$\mathbf{g}_t \gets \operatorname{sign}\left(\nabla_{\mathbf{x}} \mathcal{L}\left(\boldsymbol{f}_{\theta}\left(\boldsymbol{x}_{t-1}^* \right), \mathbf{y}\right)\right)$;

        $\boldsymbol{\delta}_t= \epsilon \mathbf{g}_t$;

        $\boldsymbol{x}_{am}^* = \boldsymbol{x}_{t-1}^* + \gamma \boldsymbol{\delta}_t$;
		
		$\lambda_t \gets {f^k_{\theta}}( \boldsymbol{x}_{t-1}^* + \boldsymbol{\delta}_t) $    $\qquad$  where  $ \boldsymbol{y}[k] = 1$;
		
		$\boldsymbol{x}_{train} \gets \lambda_t \boldsymbol{x}_{t-1}^* + (1-\lambda_t) \boldsymbol{x}_{am}^* $;
		
		$\boldsymbol{x}_{t}^* \gets \boldsymbol{x}_{t-1}^* + \epsilon \mathbf{g}_t$;
		
		$\theta_{t+1} \leftarrow \theta_t + \alpha \frac{1}{|\mathbb{B}|}\sum\limits^{\mathbb{B}} \nabla \mathcal{L}_{PCO}(\boldsymbol{f}_{\theta}(\boldsymbol{x}_{train}),\boldsymbol{y})$;
		}
	}
	
	return ${\theta}$
\end{algorithm}

\subsection{Mitigating Catastrophic Overfitting in FAT}
Current FAT methods fall short of reversing the trend of catastrophic overfitting. Once the inner-level optimization process collapses, the algorithm can only progress towards overfitting. However, in FGSM-PCO, training examples are generated by a fusion method. When catastrophic overfitting occurs, this adaptive fusion mechanism ensures retain more of the previous AEs, which avoids the reliance on current ineffective AEs as shown in Fig.\ref{bilevel}. Moreover, the inner optimization problem is closely related to the model parameters. After updating these parameters, the inner layer optimization problem can also escape from being trapped in the current overtting solution. In contrast, other FAT methods solely focus on averting models from falling into catastrophic overfitting without addressing how to rectify this trend once it occurs.

\begin{figure}[htbp]
	\centering
	\begin{subfigure}[b]{0.47\textwidth}
		\includegraphics[width=\linewidth]{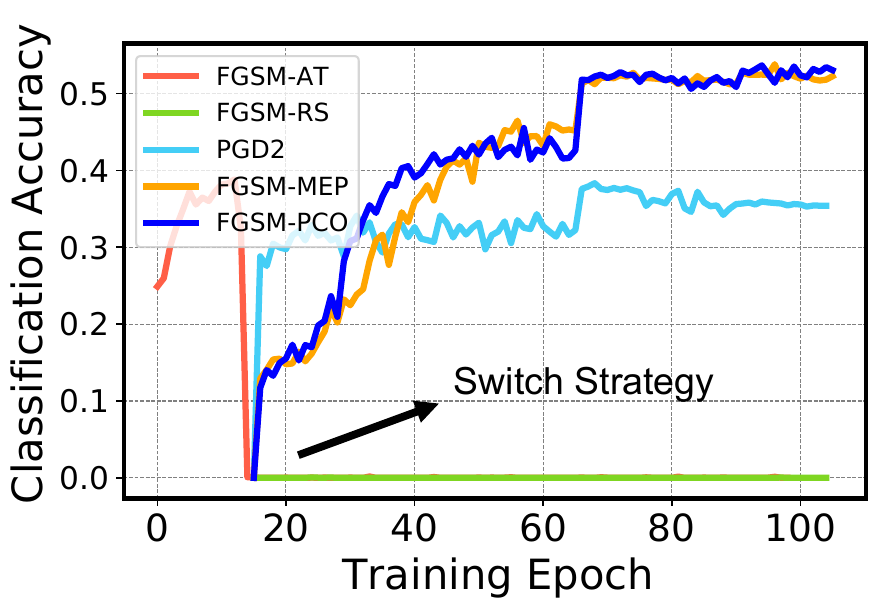}
		\caption{CIFAR10}
            \label{conduct_FGSM}
	\end{subfigure}
	\hfill
	\centering 
	\begin{subfigure}[b]{0.48\textwidth}
		\includegraphics[width=\linewidth]{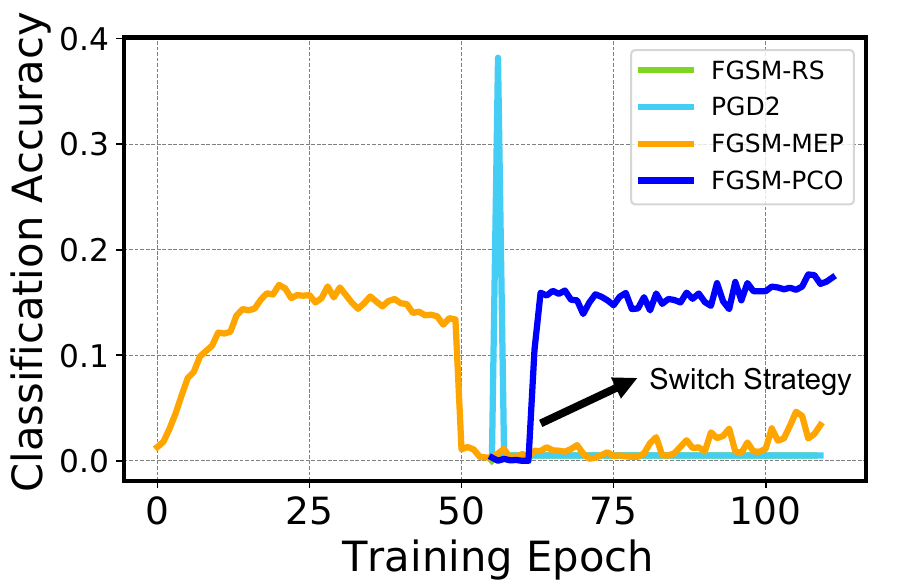}
		\caption{Tiny-ImageNet}
		\label{conduct_MEP}
	\end{subfigure}
	\caption{The performance of different FAT methods when the FGSM-AT and FGSM-MEP occur catastrophic overfitting.}
	\label{correct_ability}
\end{figure}

To validate this capability, we use FGSM-AT and FGSM-MEP to train ResNet18 \cite{KaimingHe2016IdentityMI} on CIFAR10 \cite{AlexKrizhevsky2009LearningML} and PreActResNet18 \cite{KaimingHe2016IdentityMI} on Tiny-ImageNet \cite{deng2009imagenet}, respectively. We switch to other FAT strategies when the model occurs catastrophic overfitting. Fig.~\ref{correct_ability} shows FGSM-AT occurs overfitting at $16-th$ training epoch and FGSM-MEP occurs at $50-th$ training epoch. When the inner optimization collapses, in both cases, our method FGSM-PCO can effectively correct the catastrophic overfitting issue.

In summary, the proposed framework with tailored training loss, can prevent the collapse of the inner optimization problem, thereby preventing the catastrophic overfitting phenomenon in FAT. Meanwhile, the adaptive fusion mechanism corrects the potential overfitting trend and ensures the efficacy of training.

\section{Experimental Results}
In this section, we present experimental results to evaluate the effectiveness of FGSM-PCO in comparison to other AT methods. After that, we conduct an ablation analysis to highlight the contributions of the proposed components. Experiments of CIFAR datasets are conducted on an NVIDIA RTX 4090 GPU, and experiments of Tiny-ImageNet are on an NVIDIA V100 GPU. The code will be released at https://github.com/HandingWangXDGroup/FGSM-PCO.

\subsection{Experimental Settings}
\subsubsection{Datasets and Models}
We evaluate the performance of our method on CIFAR10, CIFAR100 \cite{AlexKrizhevsky2009LearningML} and Tiny-ImageNet \cite{deng2009imagenet} datasets. The CIFAR10 dataset consists of 50000 images across 10 classes, with 5000 images per class, while CIFAR100 contains 100 classes with 500 images each. The size of each image in these two datasets is 32 $\times$ 32 $\times$ 3. Tiny-ImageNet contains 100000 images of 200 classes (500 for each class), and each image is 64 $\times$ 64 $\times$ 3. Our base models are ResNet18 \cite{KaimingHe2016IdentityMI}, PreActResNet18 \cite{KaimingHe2016IdentityMI}, and WideResNet34-10 \cite{SergeyZagoruyko2016WideRN}.

\subsubsection{Compared Methods}
We choose the PGD-AT \cite{madry2017towards} and FGSM-AT \cite{goodfellow2014explaining} as baselines for multi-step and single-step guided AT methods. In addition, we compare our FGSM-PCO method with six state-of-the-art AT methods, including FGSM-RS \cite{EricWong2020FastIB}, FGSM-GA \cite{MaksymAndriushchenko2020UnderstandingAI}, FreeAT \cite{EricWong2020FastIB}, TRADES \cite{zhang2019theoretically}, and FGSM-PGI \cite{XiaojunJia2022PriorGuidedAI}, which includes FGSM-EP, and FGSM-MEP.
\subsubsection{Training Details}
For all compared methods, the maximum perturbation is set to $8/255$. For FGSM-based AT methods, the attack step size to $8/255$, while for PGD-based AT methods, the step size is $2/255$ over 10 iterations. We utilize the SGD optimizer with a momentum of 0.9 and a weight decay of $5 \times 10^{-4}$. The initial learning rate is set to 0.1, following the settings in \cite{rice2020overfitting, pang2020bag, XiaojunJia2022PriorGuidedAI}. The training is conducted for 110 epochs, with the learning rate being divided by 10 at the 100\textsuperscript{th} and 105\textsuperscript{th} epochs. For our FGSM-PCO parameters, the regularization coefficient $\beta$ is set to 10, and the amplification parameter $\gamma$ is set to 2.
\subsubsection{Evaluation Metrics}
We evaluate the classification accuracy of all the AT models under the FGSM \cite{goodfellow2014explaining}, PGD \cite{madry2017towards}, C$\&$W \cite{carlini2017towards}, APGD \cite{croce2020reliable}, FAB \cite{croce2020minimally}, Square \cite{andriushchenko2020square} and AutoAttack (AA) \cite{croce2020reliable}, where PGD attack includes PGD10, PGD20 and PGD50 three version, APGD uses the DLR loss version. We report the result of the last checkpoint and the best checkpoint under the PGD10 attack. In addition, we also evaluate the classification accuracy of the clean examples to show the generalization of our method.

\begin{table}[htbp]
  \centering
  \caption{Accuracy (\%) and training time (min) of compared AT models on ResNet18 with the CIFAR10 dataset. The number in bold indicates the best.}
  \resizebox{\textwidth}{3.3cm}{
    \begin{tabular}{cccccccccccc}
    \toprule
    \toprule
    Method &       & Clean Acc & FGSM  & PGD10 & PGD20 & PGD50 & C\&W  & APGD  & Square & AA & Time \\
    \midrule
    \multirow{2}[2]{*}{PGD-AT} & Best  & 82.57  & 63.93  & 53.19  & 52.42  & 52.21  & 48.01  & 51.22  & 55.70  & 48.77  & \multirow{2}[2]{*}{199} \\
          & Last  & 82.99  & 64.20  & 53.05  & 52.14  & 51.96  & 47.51  & 50.66  & 55.23  & 48.23  &  \\
    \midrule
    \multirow{2}[2]{*}{TRADES} & Best  & 82.03  & 64.28  & 54.06  & 53.35  & 53.16  & 46.22  & 50.66  & 55.83  & 49.47  & \multirow{2}[2]{*}{241} \\
          & Last  & 81.80  & 63.57  & 53.89  & 53.23  & 53.05  & 46.31  & 50.39  & 55.54  & 49.56  &  \\
    \midrule
    \multirow{2}[2]{*}{FGSM-RS} & Best  & 72.95  & 52.64  & 41.40  & 40.57  & 40.40  & 36.83  & 39.95  & 44.75  & 37.67  & \multirow{2}[2]{*}{40} \\
          & Last  & 84.24  & 0.00  & 0.00  & 0.00  & 0.00  & 0.00  & 0.00  & 0.00  & 0.00  &  \\
    \midrule
    \multirow{2}[2]{*}{FGSM-GA} & Best  & 81.27  & 59.52  & 45.49  & 43.84  & 43.52  & 42.84  & 43.15  & 40.96  & 39.29  & \multirow{2}[2]{*}{132} \\
          & Last  & \textbf{86.40} & 0.00  & 0.00  & 0.00  & 0.00  & 0.00  & 0.00  & 0.00  & 0.00  &  \\
    \midrule
    \multirow{2}[2]{*}{Free-AT} & Best  & 80.64  & 57.62  & 44.10  & 42.99  & 42.73  & 39.88  & 42.83  & 48.82  & 41.17  & \multirow{2}[2]{*}{140} \\
          & Last  & 79.88  & 55.40  & 42.00  & 40.49  & 40.18  & 38.83  & 40.37  & 46.03  & 40.34  &  \\
    \midrule
    \multirow{2}[2]{*}{FGSM-EP} & Best  & 79.28  & 65.04  & 54.34  & 53.52  & 53.54  & 45.84  & 47.85  & 52.42  & 45.66  & \multirow{2}[2]{*}{57} \\
          & Last  & 80.91  & \textbf{73.83} & 40.89  & 37.76  & 36.05  & 38.46  & 28.41  & 45.29  & 25.16  &  \\
    \midrule
    \multirow{2}[2]{*}{FGSM-MEP} & Best  & 81.72  & 64.71  & 55.13  & 54.45  & 54.29  & 47.05  & \textbf{50.39} & 55.47  & \textbf{48.23} & \multirow{2}[2]{*}{57} \\
          & Last  & 81.72  & 64.71  & 55.13  & 54.45  & 54.29  & 47.05  & \textbf{50.39} & 55.47  & \textbf{48.23} &  \\
    \midrule
    \multirow{2}[2]{*}{Ours} & Best  & \textbf{82.05} & \textbf{65.53} & \textbf{56.32} & \textbf{55.66} & \textbf{55.67} & \textbf{47.12} & 50.05  & \textbf{55.59} & 48.04  & \multirow{2}[2]{*}{60} \\
          & Last  & 82.05  & 65.53  & \textbf{56.32} & \textbf{55.66} & \textbf{55.67} & \textbf{47.12} & 50.05  & \textbf{55.59} & 48.04  &  \\
    \bottomrule
    \bottomrule
    \end{tabular}%
    }
  \label{CIFAR10}%
\end{table}%

\subsection{Comparative Results}
\subsubsection{Results on CIFAR10 and CIFAR100}
To demonstrate the effectiveness of our proposed method, we compared our method with other AT methods on ResNet18 with the CIFAR datasets. The comparison results on CIFAR10 are presented in Table \ref{CIFAR10}. For the catastrophic overfitting models, we do not report the classification accuracy under the FGSM attack because those models are overfitting to the FGSM attack, which provides a false sense of security. Our method FGSM-PCO achieves 56.32\% classification accuracy under PGD10 attack, which is 3.1\% higher than the multi-step guided AT baseline PGD-AT and 1.2\% higher than the state-of-the-art FGSM-based AT method FGSM-MEP \cite{XiaojunJia2022PriorGuidedAI}. Compared to the FGSM-based methods baseline FGSM-RS, our method achieves a 14\% improvement. It is worth mentioning that, our model achieves the best classification accuracy at the last training epoch, indicating our method can effectively solve the catastrophic overfitting problem. Our method improves a little training cost compared with the FGSM-MEP but has an improvement in both adversarial and clean examples. Although our method incurs a higher computational cost than FGSM-MEP, it saves memory on computational devices, requiring only two-thirds of the memory compared to FGSM-MEP.

\begin{table}[htbp]
  \centering
  \caption{Accuracy (\%) and training time (min) of compared AT models on WideResNet34-10 with the CIFAR100 dataset. The number in bold indicates the best.} 
    \resizebox{\textwidth}{2.5cm}{
    \begin{tabular}{cccccccccccc}
    \toprule
    \toprule
    Method &       & Clean Acc & FGSM  & PGD10 & PGD20 & PGD50 & C\&W  & APGD  & Square & AA & Time \\
    \midrule
    \multirow{2}[2]{*}{PGD-AT} & Best  & 62.45  & 41.24  & 32.36  & 31.66  & 31.41  & 27.78  & 30.56  & 34.20  & 27.64  & \multirow{2}[2]{*}{1397} \\
          & Last  & 62.46  & 40.66  & 30.97  & 30.31  & 30.02  & 27.39  & 30.39  & 33.54  & 27.40  &  \\
    \midrule
    \multirow{2}[2]{*}{TRADES} & Best  & 61.23  & 40.46  & 32.14  & 31.58  & 31.56  & 25.85  & 28.87  & 32.62  & 27.60  & \multirow{2}[2]{*}{1692} \\
          & Last  & 61.09  & 40.26  & 31.89  & 31.56  & 31.57  & 25.73  & 28.72  & 32.40  & 27.56  &  \\
    \midrule
    \midrule
    \multirow{2}[1]{*}{FGSM-RS} & Best  & 51.27  & 30.92  & 22.95  & 22.41  & 21.55  & 23.74  & 25.91  & 27.11  & 16.78  & \multirow{2}[1]{*}{281} \\
          & Last  & 63.11  & -     & 0.00  & 0.00  & 0.00  & 0.00  & 0.00  & 0.00  & 0.00  &  \\
    \midrule
    \multirow{2}[1]{*}{FGSM-MEP} & Best  & 43.42  & 28.92  & 23.77  & 23.47  & 23.53  & 18.28  & 20.39  & 22.34  & 18.34  & \multirow{2}[1]{*}{407} \\
          & Last  & \textbf{72.96 } & \textbf{65.65 } & 18.99  & 13.86  & 9.63  & 12.82  & 3.62  & 12.19  & 1.49  &  \\
    \midrule
    \multirow{2}[2]{*}{FGSM-PCO} & Best  & \textbf{65.80} & \textbf{40.41} & \textbf{29.80} & \textbf{28.71} & \textbf{28.61} & \textbf{24.91} & \textbf{27.45} & \textbf{31.94} & \textbf{24.96}  & \multirow{2}[2]{*}{421} \\
          & Last  & 65.38  & 40.40  & \textbf{29.11} & \textbf{28.26} & \textbf{28.15} & \textbf{25.51} & \textbf{27.37} & \textbf{31.37} & \textbf{24.20}  &  \\
    \bottomrule
    \bottomrule
    \end{tabular}%
    }
  \label{CIFAR100}%
\end{table}%

\begin{table}[htbp]
  \centering
  \caption{Numbers of catastrophic overfitting. 10 independent repeated experiments on the WideResNet34-10 with the CIFAR100 dataset.}
  \resizebox{\textwidth}{!}{
    \begin{tabular}{c|ccccc}
    \toprule
    \toprule
    Method & FGSM-AT & FGSM-RS & FGSM-GA & FGSM-MEP & FGSM-PCO \\
    \midrule
    Numbers of overfitting  & 10/10  & 10/10  & 9/10  & 6/10  & \textbf{0/10}  \\
    \bottomrule
    \bottomrule
    \end{tabular}%
  \label{overfitting_sum}%
  }
\end{table}%


On the CIFAR100 dataset with WideResNet34-10, we achieve the best performance both on adversarial and clean examples as shown in Table \ref{CIFAR100}. Our method achieves 29.80\% under PGD10 attack and 65.80\% classification accuracy for clean examples at the best checkpoints. Even on the last training checkpoint, our method achieves 29.11\% under PGD10 attack which is higher than the state-of-the-art FGSM-based AT method FGSM-MEP. Our method significantly improves the classification accuracy on clean examples. On the last training checkpoint, we achieve 65.38\% accuracy, which is 7\% lower than the FGSM-MEP. That is because FGSM-MEP occurs overfitting phenomenon, resulting in an ineffective perturbation. To avoid the randomness of experimental outcomes, we conduct 10 independent repeated experiments on the WideResNet34-10 with the CIFAR100 dataset and record the occurrences of catastrophic overfitting as shown in Table \ref{overfitting_sum}. All the FAT algorithms occurs catastrophic overfitting problems except FGSM-PCO. More experimental results are presented in the Supplementary Material.


\begin{table}[htbp]
  \centering
  \caption{Classification accuracy (\%) on the PreActResNet with Tiny-ImageNet dataset. The number in bold indicates the best.}
  \resizebox{0.9\textwidth}{!}{
    \begin{tabular}{cccccccc}
    \toprule
    \toprule
    Method &       & Clean Acc & FGSM  & PGD10 & PGD20 & PGD50 & Training Time \\
    \midrule
    \multirow{2}[2]{*}{PGD-AT} & Best  & 33.99 & 19.56 & 15.35 & 15.29 & 15.16 & \multirow{2}[2]{*}{1961.67} \\
          & Last  & 33.76 & 12.94 & 7.05  & 6.75  & 6.68  &  \\
    \midrule
    \multirow{2}[2]{*}{FGSM-GA} & Best  & 23.67 & 14.57 & 11.59 & 11.57 & 11.57 & \multirow{2}[2]{*}{731.50} \\
          & Last  & 34.85 & 9.87  & 0.00  & 0.00     & 0.00     &  \\
    \midrule
    \multirow{2}[2]{*}{FGSM-MEP} & Best  & 31.70  & 20.51 & 16.81 & 16.74 & 16.69 & \multirow{2}[2]{*}{523.23} \\
          & Last  & \textbf{46.07} & - & 3.09  & 1.98  & 1.36  &  \\
    \midrule
    \multirow{2}[2]{*}{Ours} & Best  & 34.96 & \textbf{22.32}& \textbf{18.17} & \textbf{18.10}  & \textbf{17.99} & \multirow{2}[2]{*}{686.67} \\
          & Last  & {37.67} & \textbf{20.33} & \textbf{14.55} & \textbf{14.37} & \textbf{14.28} &  \\
    \bottomrule
    \bottomrule
    \end{tabular}%
    }
  \label{tiny}%
\end{table}%



\begin{figure}[!htbp]
	\centering
	\begin{subfigure}[b]{0.48\textwidth}
		\includegraphics[width=\linewidth]{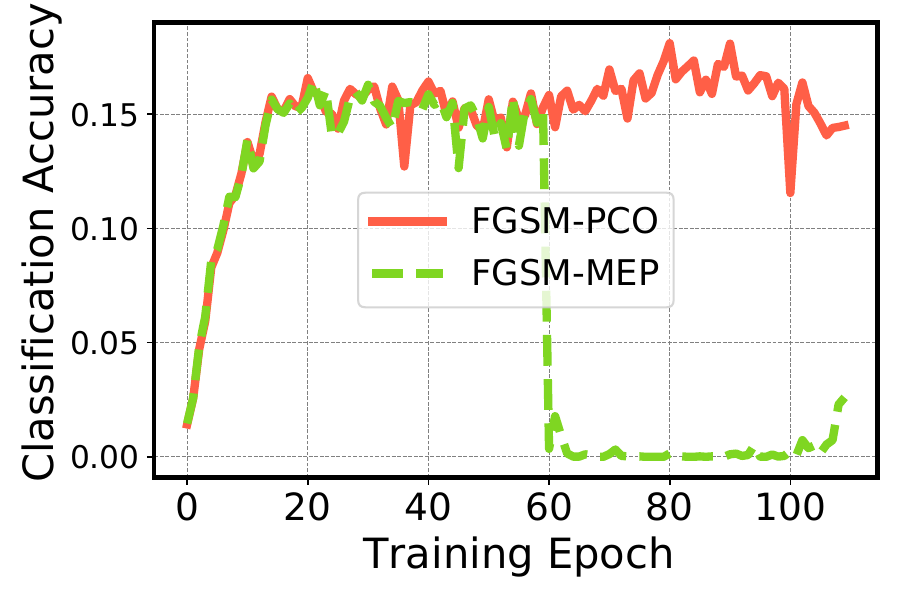}
		\caption{Adversarial Examples}
	\end{subfigure}
	\hfill
	\centering 
	\begin{subfigure}[b]{0.48\textwidth}
		\includegraphics[width=\linewidth]{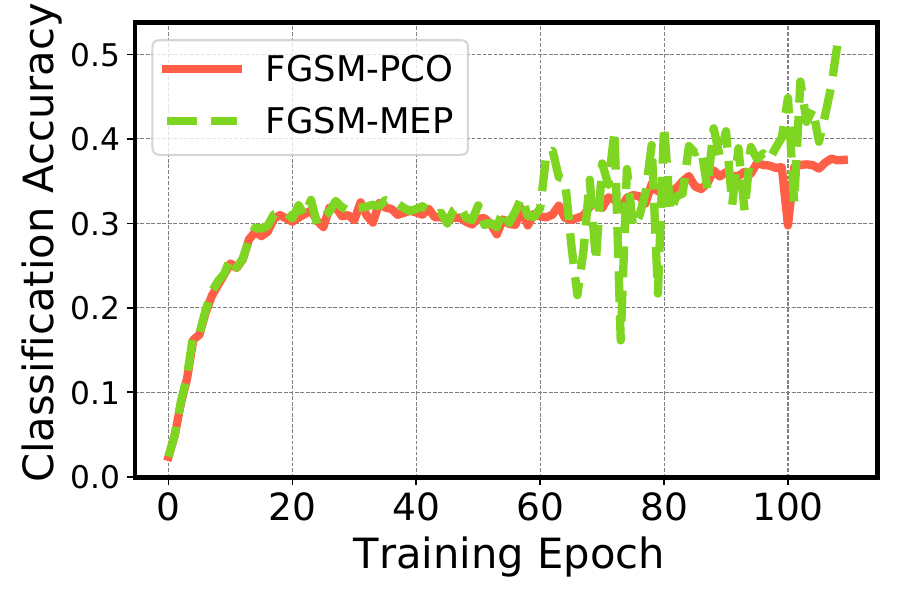}
		\caption{Clean Examples}
	\end{subfigure}
	\caption{The classification accuracy of PreActResNet18 with the Tiny-ImageNet dataset. The left figure shows the classification accuracy for AEs under PGD10 attack, and the right figure shows the accuracy for clean examples.}
	\label{tiny-pre}
\end{figure}

\subsubsection{Results on Tiny-ImageNet}
To show the performance comprehensively and further invest in the catastrophic overfitting phenomenon, we test FGSM-PCO on PreActResNet18 with a more challenging dataset Tiny-ImageNet. We use the cyclic learning rate strategy \cite{smith2017cyclical} with a maximum learning rate of 0.2. The experimental results are shown in Table \ref{tiny}. We do not report the FGSM accuracy of the overfitting models. Although the classification accuracy of our method has a gap between the best and the last model under the PGD-10 attack, FGSM-PCO still achieves the best performance. FGSM-PCO obtains 18.17\% accuracy under the PGD10 attack and 34.96\% classification accuracy on clean examples. Although FGSM-MEP utilizes the historical information to guide the initialization of the current perturbation, it occurs catastrophic overfitting shown in Fig.~\ref{tiny-pre}. Meanwhile, the classification on clean examples occurs in fluctuations. FGSM-MEP obtains 16.48\% accuracy on AEs and 32.15\% on clean examples, which is lower than our proposed method FGSM-PCO.

\begin{figure}[htbp]
	\centering
	\begin{subfigure}[b]{0.48\textwidth}
		\includegraphics[width=\linewidth]{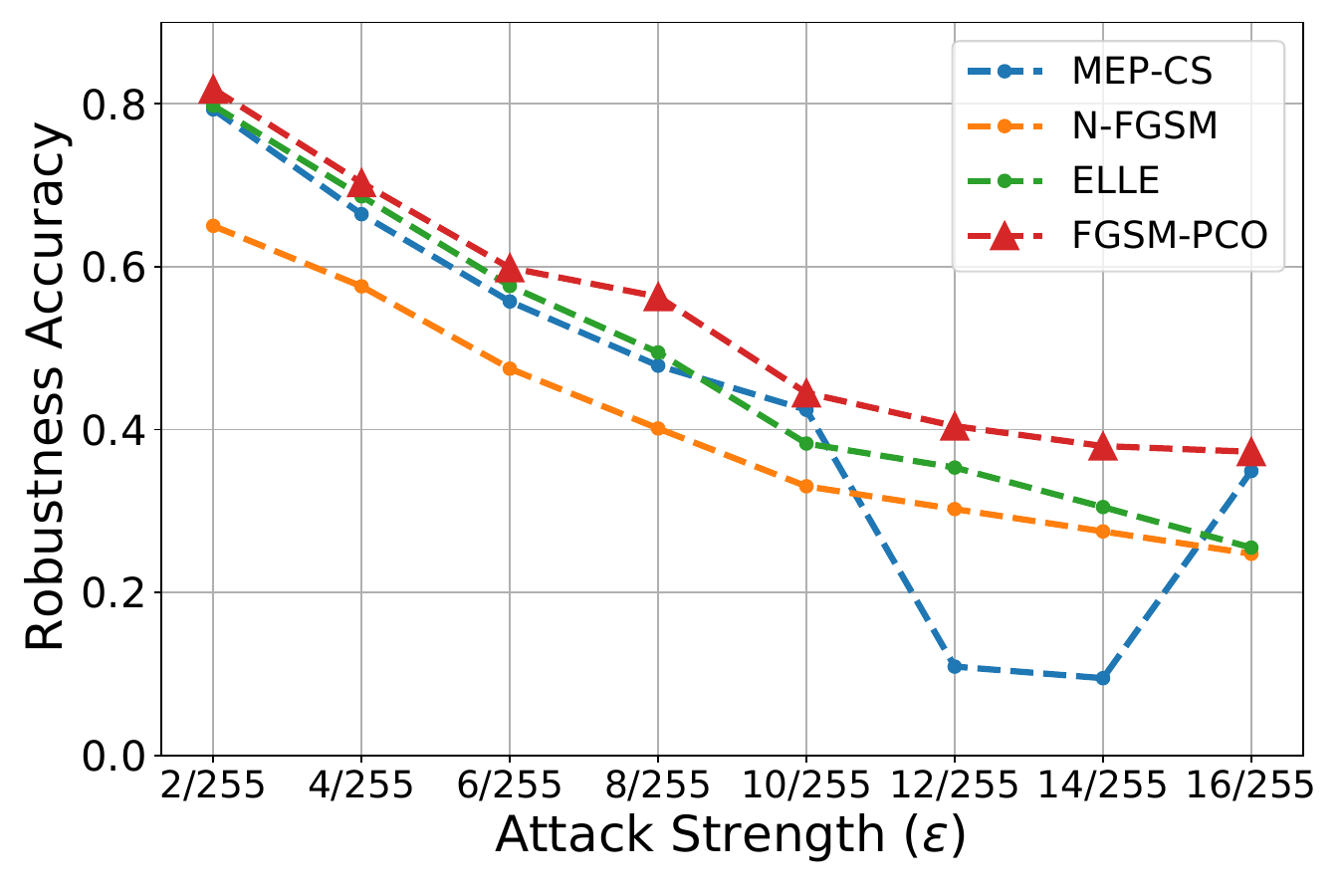}
		\caption{}
            \label{attack_strength}
	\end{subfigure}
	\hfill
	\centering 
	\begin{subfigure}[b]{0.48\textwidth}
		\includegraphics[width=\linewidth]{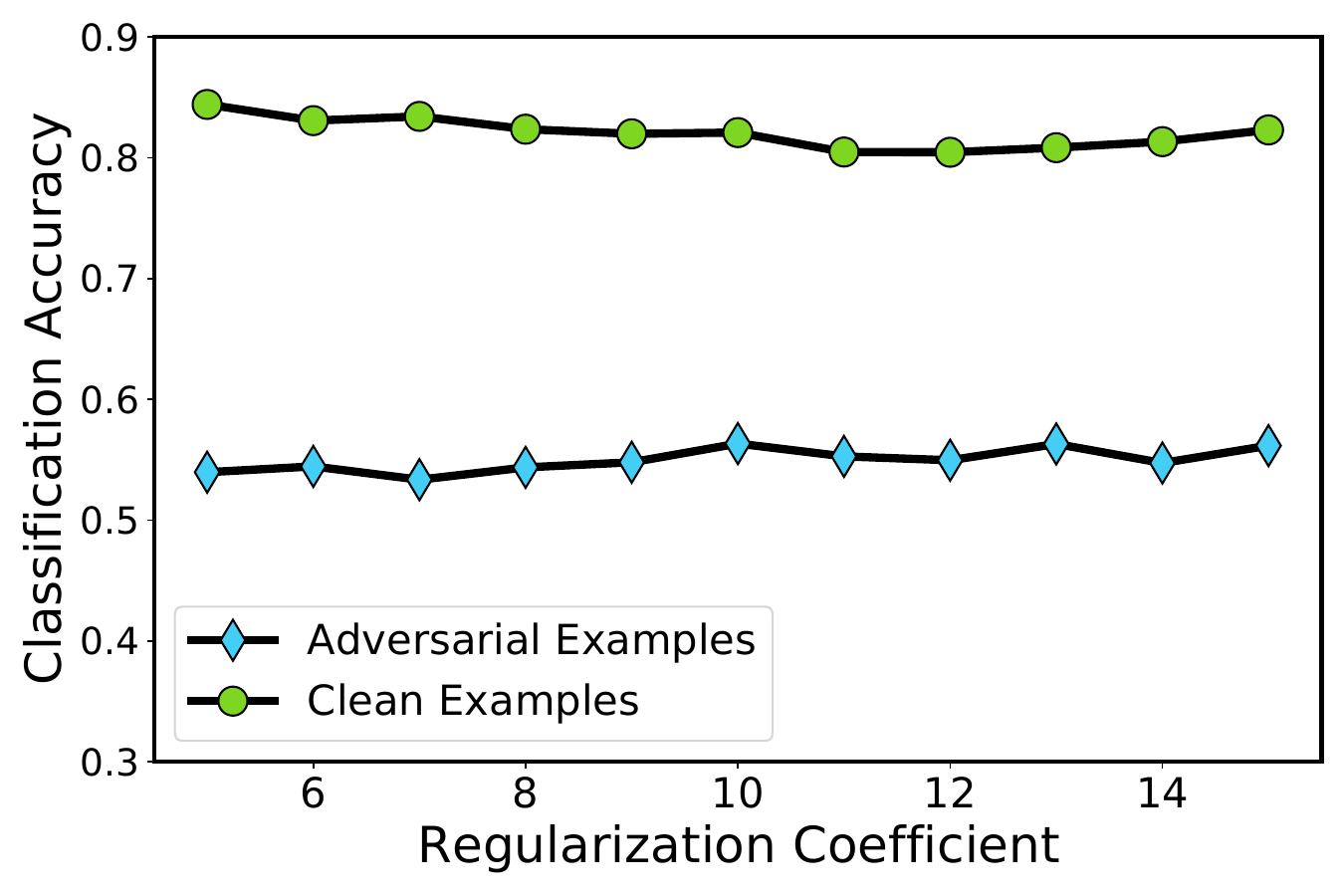}
		\caption{}
		\label{sensitive}
	\end{subfigure}
	\caption{\ref{attack_strength} shows the classification accuracy under different attack strengths. \ref{sensitive} represents the sensitivity of $\beta$ on classification for adversarial and clean examples.}
\end{figure}

\subsubsection{Results under Different Attack Strength}
To further show the effectiveness of our proposed method, we evaluate the classification accuracy under different levels of attack budgets. We compared FGSM-PCO with three methods, N-FGSM \cite{de2022make}, MEP-CS \cite{zhao2023fast} and GAT-ELLE \cite{rocamora2024efficient}, which perform well under various attack strengths. The results are shown in Fig.~\ref{attack_strength}.

\subsubsection{Sensitivity Analysis}
We investigate the regularization parameter $\beta$ in Eq.(\ref{regularization}) on ResNet18 with the CIFAR10 dataset. We present the accuracy on adversarial and clean examples as shown in Fig.~\ref{sensitive}. The regularization parameter is set to different values $\beta \in [5,15]$. The sensitivity experiments show that our method achieves good stability and robustness across different choices of $\beta$.


\subsection{Ablation Study}
To better understand the effects of each component and highlight our contributions, we conduct the ablation experiments on ResNet18 with the CIFAR10 dataset.

\begin{table}[htbp]
  \centering
  \caption{Ablation study of the proposed method.}
  \begin{threeparttable}
  \resizebox{0.75\textwidth}{!}{
    \begin{tabular}{c|c|c|c|c|c|c|c}
    \toprule
    \toprule
    \multirow{3}[2]{*}{Component} & Loss\tnote{\dag }  & \XSolidBrush     & \XSolidBrush     & \Checkmark     & \Checkmark     & \XSolidBrush    & \Checkmark \\
          & {Fusion\tnote{\ddag}} & \XSolidBrush     & \Checkmark     & \Checkmark     & \XSolidBrush    & \Checkmark     & \Checkmark \\
          & {Adaptive\tnote{*} }  & \XSolidBrush     & \XSolidBrush     & \XSolidBrush     & \Checkmark     & \Checkmark     & \Checkmark \\
    \midrule
    \multirow{2}[2]{*}{Clean Acc} & Best  & 87.44 & 76.40 & 82.01 & 80.01 & 88.88 & 82.48 \\
          & Last  & 87.90 & 89.81 & 81.97 & 80.30 & 88.65 & 82.48 \\
    \midrule
    \multirow{2}[2]{*}{PGD10} & Best  & 48.74 & 39.91 & 54.27 & 49.04 & 50.67 & \textbf{56.12} \\
          & Last  & 48.42 & 10.18 & 53.56 & 48.81 & 50.57 & \textbf{56.12} \\
    \bottomrule
    \bottomrule
    \end{tabular}%
    }
    	\begin{tablenotes}
			\item[\dag] represents the proposed regularization.
			\item[\ddag] represents the perturbation fusion strategy.
			\item[*] represents the adaptive mechanism.
		\end{tablenotes}
	\end{threeparttable}
  \label{ablation}%
\end{table}%

\subsubsection{Effect of Each Component}
For the proposed loss function, AEs fusion strategy and adaptive fusion factor, we conduct experiments on ResNet18 on the CIFAR10 dataset and the results are presented in Table \ref{ablation}. The result shows that our proposed regularization can effectively prevent catastrophic overfitting and well guide the proposed AT method, achieving the best performance when the AT algorithm contains the three components.

\subsubsection{Fusion strategies}
\begin{figure}[htbp]
	\centering
	\begin{subfigure}[b]{0.49\textwidth}
		\includegraphics[width=\linewidth]{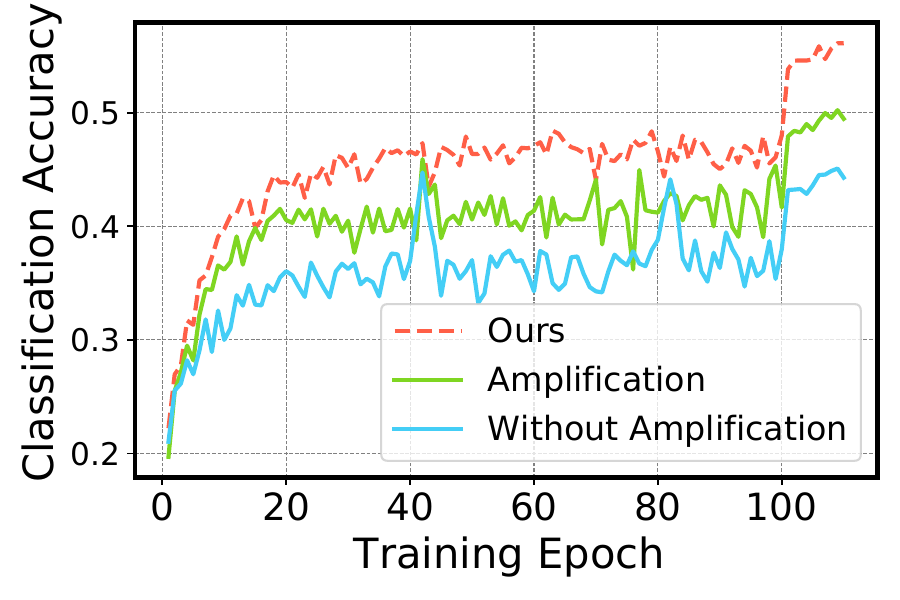}
		\caption{Adversarial Examples}
	\end{subfigure}
	\hfill
	\centering 
	\begin{subfigure}[b]{0.49\textwidth}
		\includegraphics[width=\linewidth]{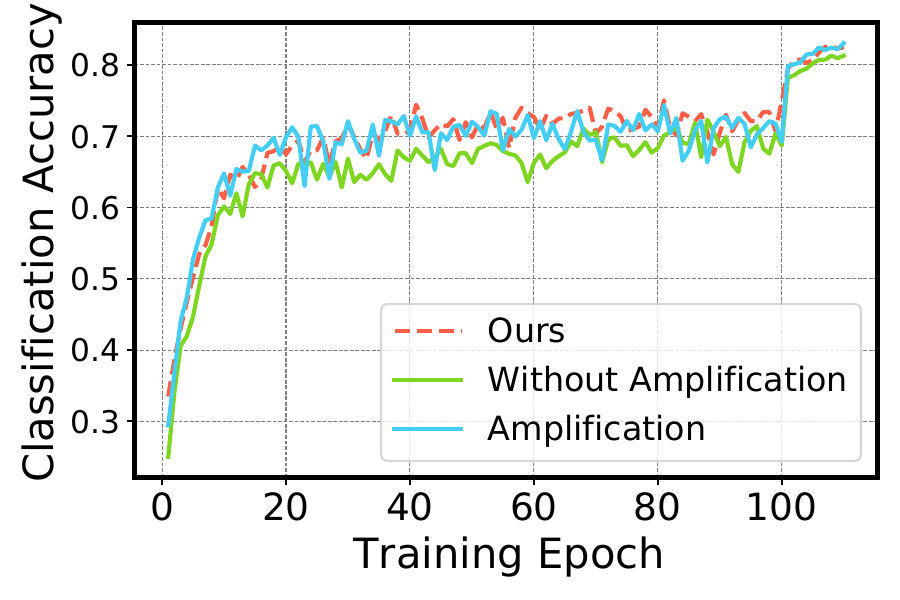}
		\caption{Clean Examples}
	\end{subfigure}
	\caption{The classification accuracy of different fusion strategies on ResNet18 with the CIFAR10. The left figure shows the classification accuracy for AEs under PGD10 attack, and the right figure shows the accuracy for clean examples.}
	\label{mixup_figure}
\end{figure}

To demonstrate that the enhanced classification accuracy does not result from the generalization introduced by the mixup technology. We conduct experiments on ResNet18 with the CIFAR10 dataset, and three fusion strategies are compared. The first is our fusion method using the previous and current perturbation with an amplified parameter. The second is to fuse the amplified current AEs with clean examples (Amplification), and the third one is to fuse the current AEs and clean examples without amplification (Without Amplification). The results shown in Fig.~\ref{mixup_figure}, our proposed method achieves the best performance among the three strategies without decreasing the classification accuracy for clean examples.

\section{Conclusions}
In this study, we demonstrate that avoiding the collapse of the inner optimization problem can effectively prevent catastrophic overfitting in FAT. Our proposed method FGSM-PCO, fuses the previous AEs and an amplification AEs with an adaptive mechanism, which can correct the catastrophic overfitting trend. Our method is compared with the seven state-of-the-art AT methods on three datasets and three models, and the experimental results demonstrate the effectiveness of our method.

Although our method achieves good performance on the chosen datasets, it also has a limitation. The fusion coefficient of our method is decided according to the classification confidence of the AEs, which may become less effective when the method faces a vast number of categories in an open-world scenario. In the future, we plan to explore a more efficient connection between previous and current training epochs.

\section*{Acknowledgements}
This work was supported in part by the National Natural Science Foundation of China (No. 62376202, 62192734) and the Innovation Fund of Xidian University (YJSJ24012).

\bibliographystyle{splncs04}

\begin{thebibliography}{10}
\providecommand{\url}[1]{\texttt{#1}}
\providecommand{\urlprefix}{URL }
\providecommand{\doi}[1]{https://doi.org/#1}

\bibitem{andriushchenko2020square}
Andriushchenko, M., Croce, F., Flammarion, N., Hein, M.: Square attack: a query-efficient black-box adversarial attack via random search. In: European Conference on Computer Vision. pp. 484--501. Springer (2020)

\bibitem{MaksymAndriushchenko2020UnderstandingAI}
Andriushchenko, M., Flammarion, N.: Understanding and improving fast adversarial training. Advances in Neural Information Processing Systems  \textbf{33},  16048--16059 (2020)

\bibitem{brendel2017decision}
Brendel, W., Rauber, J., Bethge, M.: Decision-based adversarial attacks: Reliable attacks against black-box machine learning models. In: International Conference on Learning Representations (2018)

\bibitem{carlini2017towards}
Carlini, N., Wagner, D.: Towards evaluating the robustness of neural networks. In: 2017 IEEE Symposium on Security and Privacy (SP). IEEE (2017)

\bibitem{chakraborty2018adversarial}
Chakraborty, A., Alam, M., Dey, V., Chattopadhyay, A., Mukhopadhyay, D.: Adversarial attacks and defences: A survey. arXiv preprint arXiv:1810.00069  (2018)

\bibitem{chen2017zoo}
Chen, P.Y., Zhang, H., Sharma, Y., Yi, J., Hsieh, C.J.: Zoo: Zeroth order optimization based black-box attacks to deep neural networks without training substitute models. In: Proceedings of the 10th ACM workshop on artificial intelligence and security. pp. 15--26 (2017)

\bibitem{colson2007overview}
Colson, B., Marcotte, P., Savard, G.: An overview of bilevel optimization. Annals of operations research  \textbf{153},  235--256 (2007)

\bibitem{croce2020minimally}
Croce, F., Hein, M.: Minimally distorted adversarial examples with a fast adaptive boundary attack. In: International Conference on Machine Learning. pp. 2196--2205. PMLR (2020)

\bibitem{croce2020reliable}
Croce, F., Hein, M.: Reliable evaluation of adversarial robustness with an ensemble of diverse parameter-free attacks. In: International conference on machine learning. pp. 2206--2216. PMLR (2020)

\bibitem{dempe2015bilevel}
Dempe, S., Kalashnikov, V., P{\'e}rez-Vald{\'e}s, G.A., Kalashnykova, N.: Bilevel programming problems. Energy Systems. Springer, Berlin  \textbf{10},  978--3 (2015)

\bibitem{deng2009imagenet}
Deng, J., Dong, W., Socher, R., Li, L.J., Li, K., Fei-Fei, L.: Imagenet: A large-scale hierarchical image database. In: 2009 IEEE conference on computer vision and pattern recognition. pp. 248--255. Ieee (2009)

\bibitem{duan2021adversarial}
Duan, R., Mao, X., Qin, A.K., Chen, Y., Ye, S., He, Y., Yang, Y.: Adversarial laser beam: Effective physical-world attack to dnns in a blink. In: Proceedings of the IEEE/CVF Conference on Computer Vision and Pattern Recognition. pp. 16062--16071 (2021)

\bibitem{goodfellow2014explaining}
Goodfellow, I.J., Shlens, J., Szegedy, C.: Explaining and harnessing adversarial examples. arXiv preprint arXiv:1412.6572  (2014)

\bibitem{gu2017badnets}
Gu, T., Dolan-Gavitt, B., Garg, S.: Badnets: Identifying vulnerabilities in the machine learning model supply chain. arXiv preprint arXiv:1708.06733  (2017)

\bibitem{KaimingHe2016IdentityMI}
He, K., Zhang, X., Ren, S., Sun, J.: Identity mappings in deep residual networks. In: Computer Vision--ECCV 2016: 14th European Conference, Amsterdam, The Netherlands, October 11--14, 2016, Proceedings, Part IV 14. pp. 630--645. Springer (2016)

\bibitem{ilyas2018black}
Ilyas, A., Engstrom, L., Athalye, A., Lin, J.: Black-box adversarial attacks with limited queries and information. In: International conference on machine learning. pp. 2137--2146. PMLR (2018)

\bibitem{XiaojunJia2022PriorGuidedAI}
Jia, X., Zhang, Y., Wei, X., Wu, B., Ma, K., Wang, J., Cao, X.: Prior-guided adversarial initialization for fast adversarial training. In: European Conference on Computer Vision. pp. 567--584. Springer (2022)

\bibitem{de2022make}
de~Jorge~Aranda, P., Bibi, A., Volpi, R., Sanyal, A., Torr, P., Rogez, G., Dokania, P.: Make some noise: Reliable and efficient single-step adversarial training. Advances in Neural Information Processing Systems  \textbf{35},  12881--12893 (2022)

\bibitem{kim2021understanding}
Kim, H., Lee, W., Lee, J.: Understanding catastrophic overfitting in single-step adversarial training. In: Proceedings of the AAAI Conference on Artificial Intelligence. vol.~35, pp. 8119--8127 (2021)

\bibitem{AlexKrizhevsky2009LearningML}
Krizhevsky, A.: Learning multiple layers of features from tiny images (2009)

\bibitem{li2022backdoor}
Li, Y., Jiang, Y., Li, Z., Xia, S.T.: Backdoor learning: A survey. IEEE Transactions on Neural Networks and Learning Systems  (2022)

\bibitem{liu2020privacy}
Liu, X., Xie, L., Wang, Y., Zou, J., Xiong, J., Ying, Z., Vasilakos, A.V.: Privacy and security issues in deep learning: A survey. IEEE Access  \textbf{9},  4566--4593 (2020)

\bibitem{madry2017towards}
Madry, A., Makelov, A., Schmidt, L., Tsipras, D., Vladu, A.: Towards deep learning models resistant to adversarial attacks. In: International Conference on Learning Representations (2018)

\bibitem{ChengzhiMao2019MetricLF}
Mao, C., Zhong, Z., Yang, J., Vondrick, C., Ray, B.: Metric learning for adversarial robustness. Advances in neural information processing systems  \textbf{32} (2019)

\bibitem{moosavi2016deepfool}
Moosavi-Dezfooli, S.M., Fawzi, A., Frossard, P.: Deepfool: a simple and accurate method to fool deep neural networks. In: Proceedings of the IEEE conference on computer vision and pattern recognition. pp. 2574--2582 (2016)

\bibitem{SeyedMohsenMoosaviDezfooli2019RobustnessVC}
Moosavi-Dezfooli, S.M., Fawzi, A., Uesato, J., Frossard, P.: Robustness via curvature regularization, and vice versa. In: Proceedings of the IEEE/CVF Conference on Computer Vision and Pattern Recognition. pp. 9078--9086 (2019)

\bibitem{pang2020bag}
Pang, T., Yang, X., Dong, Y., Su, H., Zhu, J.: Bag of tricks for adversarial training. arXiv preprint arXiv:2010.00467  (2020)

\bibitem{ChongliQin2019AdversarialRT}
Qin, C., Martens, J., Gowal, S., Krishnan, D., Dvijotham, K., Fawzi, A., De, S., Stanforth, R., Kohli, P.: Adversarial robustness through local linearization. Advances in neural information processing systems  \textbf{32} (2019)

\bibitem{rice2020overfitting}
Rice, L., Wong, E., Kolter, Z.: Overfitting in adversarially robust deep learning. In: International Conference on Machine Learning. pp. 8093--8104. PMLR (2020)

\bibitem{rocamora2024efficient}
Rocamora, E.A., Liu, F., Chrysos, G., Olmos, P.M., Cevher, V.: Efficient local linearity regularization to overcome catastrophic overfitting. In: The Twelfth International Conference on Learning Representations (2024)

\bibitem{shafahi2019adversarial}
Shafahi, A., Najibi, M., Ghiasi, M.A., Xu, Z., Dickerson, J., Studer, C., Davis, L.S., Taylor, G., Goldstein, T.: Adversarial training for free! Advances in Neural Information Processing Systems  \textbf{32} (2019)

\bibitem{shokri2017membership}
Shokri, R., Stronati, M., Song, C., Shmatikov, V.: Membership inference attacks against machine learning models. In: 2017 IEEE symposium on security and privacy (SP). pp. 3--18. IEEE (2017)

\bibitem{smith2017cyclical}
Smith, L.N.: Cyclical learning rates for training neural networks. In: 2017 IEEE winter conference on applications of computer vision (WACV). pp. 464--472. IEEE (2017)

\bibitem{szegedy2013intriguing}
Szegedy, C., Zaremba, W., Sutskever, I., Bruna, J., Erhan, D., fellow, I., Fergus, R.: Intriguing properties of neural networks. arXiv preprint arXiv:1312.6199  (2013)

\bibitem{wang2019improving}
Wang, Y., Zou, D., Yi, J., Bailey, J., Ma, X., Gu, Q.: Improving adversarial robustness requires revisiting misclassified examples. In: International Conference on Learning Representations (2019)

\bibitem{wang2023better}
Wang, Z., Pang, T., Du, C., Lin, M., Liu, W., Yan, S.: Better diffusion models further improve adversarial training. arXiv preprint arXiv:2302.04638  (2023)

\bibitem{wang2023adversarial}
Wang, Z., Wang, H., Tian, C., Jin, Y.: Adversarial training of deep neural networks guided by texture and structural information. In: Proceedings of the 31st ACM International Conference on Multimedia. pp. 4958--4967 (2023)

\bibitem{EricWong2020FastIB}
Wong, E., Rice, L., Kolter, J.Z.: Fast is better than free: Revisiting adversarial training. Learning  (2020)

\bibitem{SergeyZagoruyko2016WideRN}
Zagoruyko, S., Komodakis, N.: Wide residual networks. arXiv: Computer Vision and Pattern Recognition  (2016)

\bibitem{DinghuaiZhang2019YouOP}
Zhang, D., Zhang, T., Lu, Y., Zhu, Z., Dong, B.: You only propagate once: Accelerating adversarial training via maximal principle. Advances in neural information processing systems  \textbf{32} (2019)

\bibitem{zhang2019theoretically}
Zhang, H., Yu, Y., Jiao, J., Xing, E., El~Ghaoui, L., Jordan, M.: Theoretically principled trade-off between robustness and accuracy. In: International conference on machine learning. pp. 7472--7482. PMLR (2019)

\bibitem{YihuaZhang2022RevisitingAA}
Zhang, Y., Zhang, G., Khanduri, P., Hong, M., Chang, S., Liu, S.: Revisiting and advancing fast adversarial training through the lens of bi-level optimization (2022)

\bibitem{zhao2023fast}
Zhao, M., Zhang, L., Kong, Y., Yin, B.: Fast adversarial training with smooth convergence. In: Proceedings of the IEEE/CVF International Conference on Computer Vision. pp. 4720--4729 (2023)

\bibitem{zhu2019deep}
Zhu, L., Liu, Z., Han, S.: Deep leakage from gradients. Advances in neural information processing systems  \textbf{32} (2019)

\end{thebibliography}

\end{document}


\title{Supplementary Material of Preventing Catastrophic Overfitting in Fast Adversarial Training: A Bi-level Optimization Perspective} 

\titlerunning{Abbreviated paper title}

\author{Zhaoxin~Wang\inst{1}\orcidlink{0009-0009-5860-8370} \and
Handing~Wang\inst{1}\textsuperscript{\Envelope}\orcidlink{0000-0002-4805-3780} \and
Cong~Tian\inst{2}\orcidlink{0000-0002-5429-4580} \and
Yaochu~Jin\inst{3}\orcidlink{0000-0003-1100-0631}}

\authorrunning{Z.~Wang et al.}

\institute{School of Artificial Intelligence, Xidian University, Xi'an, China \\ \email{zxwang74@163.com} \and School of Computer Science and Technology, Xidian University, Xi'an, China \\ \email{ctian@mail.xidian.edu.cn} \and School of Engineering, Westlake University, Zhejiang Hangzhou, China \\ \email{jinyaochu@westlake.edu.cn}}

\maketitle

\let\thefootnote\relax\footnotetext{\textsuperscript{\Envelope} Corresponding Author: hdwang@xidian.edu.cn}



\begin{table}[htbp]
  \centering
  \caption{Accuracy (\%) and training time (min) of compared AT models on WideResNet34-10 with the CIFAR10 dataset. The number in bold indicates the best.} 
\resizebox{\textwidth}{!}{
    \begin{tabular}{cccccccccccc}
    \toprule
    \toprule
    Method &       & Clean Acc & FGSM  & PGD10 & PGD20 & PGD50 & C\&W  & APGD  & Square & AA    & Time \\
    \midrule
    \multirow{2}[2]{*}{PGD-AT} & Best  & 87.30  & 68.92  & 55.21  & 53.96  & 53.49  & 51.80  & 54.42  & 60.40  & 51.20  & \multirow{2}[2]{*}{1397} \\
          & Last  & 87.39  & 68.20  & 54.12  & 52.85  & 52.49  & 50.68  & 53.37  & 59.36  & 50.57  &  \\
    \midrule
    \multirow{2}[2]{*}{TRADES} & Best  & 85.70  & 68.28  & 57.21  & 56.10  & 55.87  & 50.72  & 54.85  & 59.84  & 53.36  & \multirow{2}[2]{*}{1692} \\
          & Last  & 85.70  & 68.28  & 57.21  & 56.10  & 55.87  & 50.72  & 54.85  & 59.84  & 53.36  &  \\
    \midrule
    \midrule
    \multirow{2}[2]{*}{FGSM-RS} & Best  & 75.10  & 59.00  & 44.66  & 43.29  & 42.96  & 38.68  & 44.98  & 50.28  & 40.27  & \multirow{2}[2]{*}{281} \\
          & Last  & 86.19  & -     & 0.00  & 0.00  & 0.00  & 0.00  & 0.00  & 0.00  & 0.00  &  \\
    \midrule
    \multirow{2}[2]{*}{Free-AT} & Best  & 71.79  & 51.37  & 41.75  & 41.13  & 40.99  & 35.67  & 43.81  & 44.33  & 39.22  & \multirow{2}[2]{*}{969} \\
          & Last  & 71.79  & 51.37  & 41.75  & 41.13  & 40.99  & 35.67  & 43.81  & 44.33  & 39.22  &  \\
    \midrule
    \multirow{2}[2]{*}{FGSM-MEP} & Best  & 83.43  & 67.73  & \textbf{58.13} & \textbf{57.52} & \textbf{57.51} & 49.62  & 53.35  & 58.13  & 51.54  & \multirow{2}[2]{*}{407} \\
          & Last  & 85.63  & 69.09  & 57.47  & 56.48  & 56.20  & 49.82  & 52.89  & 58.45  & 51.06  &  \\
    \midrule
    \multirow{2}[2]{*}{FGSM-PCO} & Best  & \textbf{87.38} & \textbf{69.78} & 57.82  & 57.12  & 56.96  & \textbf{51.27} & \textbf{54.34} & \textbf{59.88} & \textbf{51.84} & \multirow{2}[2]{*}{421} \\
          & Last  & \textbf{87.38} & \textbf{69.78} & \textbf{57.82} & \textbf{57.12} & \textbf{56.96} & \textbf{51.27} & \textbf{54.34} & \textbf{59.88} & \textbf{51.84} &  \\
    \bottomrule
    \bottomrule
    \end{tabular}%
    }
  \label{CIFAR10-Wide}%
\end{table}%

\section{Experimental Results}
The classification accuracy of WideResNet34-10 with the CIFAR10 dataset is shown in Table \ref{CIFAR10-Wide}. On the WideResNet34-10 model, we achieve good performance, especially for clean examples, which reach 87.38\% accuracy. To comprehensively evaluate the performance of various AT methods and investigate the overfitting phenomenon, we conduct experiments with a smaller model as the backbone on datasets where catastrophic overfitting occurs. Table \ref{CIFAR100} presents the results on CIFAR100 with the ResNet18 model. The results demonstrate that FGSM-MEP effectively prevents the catastrophic overfitting problem observed in the WideResNet34-10 model. Our method, FGSM-PCO, achieves improvements both on clean examples and AEs, with only a 0.1\% lower performance than FGSM-MEP under the CW attack at the last checkpoint. It is noteworthy that our method incurs a higher computational cost than FGSM-MEP but saves memory on computational devices, requiring only two-thirds of the memory compared to FGSM-MEP.

\begin{table*}[htbp]  
	\centering  
	\caption{Accuracy (\%) and training time (min) of compared AT models on ResNet18 with the CIFAR100 dataset. The number in bold indicates the best.} 
	\resizebox{\textwidth}{!}{   
		\begin{tabular}{ccccccccccc}    
			\toprule   
			\toprule    Method &       & Clean Acc & FGSM  & PGD10 & PGD20 & PGD50 & C\&W  & APGD  & Square & Time \\    
			\midrule    
			\multirow{2}[2]{*}{PGD-AT \cite{madry2017towards}} & Best  & 58.24 & 37.84 & 29.68 & 29.20  & 29.15 & 25.09 & 27.82 & 30.73 & 
			\multirow{2}[2]{*}{191} \\          & Last  & 58.42 & 37.59 & 29.00    & 28.45 & 28.35 & 24.83 & 27.27 & 30.42 &  \\    
			\midrule    
			\multirow{2}[2]{*}{TRADES \cite{zhang2019theoretically}} & Best  & 58.38 & 37.95 & 30.53 & 30.05 & 29.95 & 23.55 & 26.28 & 30.20  & 
			\multirow{2}[2]{*}{260} \\          & Last  & 58.00    & 38.08 & 30.34 & 29.99 & 29.89 & 23.57 & 26.15 & 29.95 &  \\    
			\midrule    
			\midrule
			\multirow{2}[2]{*}{FGSM-RS \cite{EricWong2020FastIB}} & Best  & 45.64 & 28.87 & 20.89 & 20.20  & 20.24 & 17.21 & 17.82 & 20.01 & 
			\multirow{2}[2]{*}{38} \\          & Last  & 42.54 & -     & 00.00  & 00.00     & 00.00     & 00.00    & 00.00     & 00.00     &  \\    
			\midrule   
			\multirow{2}[2]{*}{FGSM-GA \cite{MaksymAndriushchenko2020UnderstandingAI}} & Best  & 46.37 & 28.56 & 21.74 & 21.43 & 21.31 & 18.18 & 19.81 & 22.21 & 
			\multirow{2}[2]{*}{137} \\          & Last  & \pmb{62.34} & -     & 00.02  & 00.00     &  00.00    &  00.00    &  00.00    & 00.00     &  \\    
			\midrule    
			\multirow{2}[2]{*}{Free-AT \cite{shafahi2019adversarial}} & Best  & 38.19 & 23.17 & 18.38 & 18.11 & 18.08 & 15.02 & 16.13 & 17.46 & 
			\multirow{2}[2]{*}{138} \\          & Last  & 38.19 & 23.17 & 18.38 & 18.11 & 18.08 & 15.02 & 16.13 & 17.46 &  \\    
			\midrule    
			\multirow{2}[2]{*}{FGSM-EP \cite{XiaojunJia2022PriorGuidedAI}} & Best  & 58.24 & 39.69 & 31.69 & 31.34 & 31.27 & 25.14 & 27.39 & 30.81 & 
			\multirow{2}[2]{*}{58} \\          & Last  & 58.20  & 39.41 & 31.39 & 30.96 & 30.92 & 24.86 & 27.28 & 30.46 &  \\    
			\midrule    
			\multirow{2}[2]{*}{FGSM-MEP \cite{XiaojunJia2022PriorGuidedAI}} & Best  & 58.79 & 39.06 & 31.83 & 31.35 & 31.35 & 25.76 & 27.88 & 31.09 & 
			\multirow{2}[2]{*}{58} \\          & Last  & 58.82 & 39.77 & 31.74 & 31.22 & 31.12 & \pmb{25.26} & \pmb{27.66} & 30.92 &  \\    
			\midrule    
			\multirow{2}[2]{*}{FGSM-HPF} & Best  & \pmb{60.20} & \pmb{39.98} & \pmb{32.39} & \pmb{31.94} & \pmb{31.85} & \pmb{25.85} & \pmb{28.16} & \pmb{31.50} & \multirow{2}[2]{*}{60} \\          & Last  & 59.80 & \pmb{39.83} & \pmb{31.89} & \pmb{31.44} & \pmb{31.36} & 25.25 & 27.62 & \pmb{31.11} &  \\    
			\bottomrule    
			\bottomrule    
		\end{tabular}%
	}
	\label{CIFAR100}%
\end{table*}%

\begin{figure}[htbp]
	\centering
	\begin{subfigure}[b]{0.48\textwidth}
		\includegraphics[width=\linewidth]{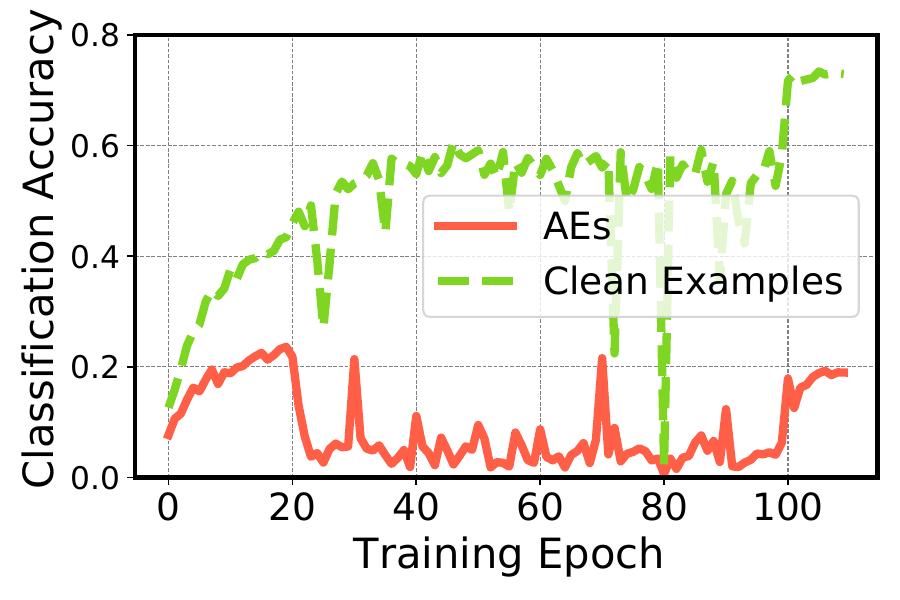}
		\caption{FGSM-MEP}
	\end{subfigure}
	\hfill
	\centering
	\begin{subfigure}[b]{0.48\textwidth}
		\includegraphics[width=\linewidth]{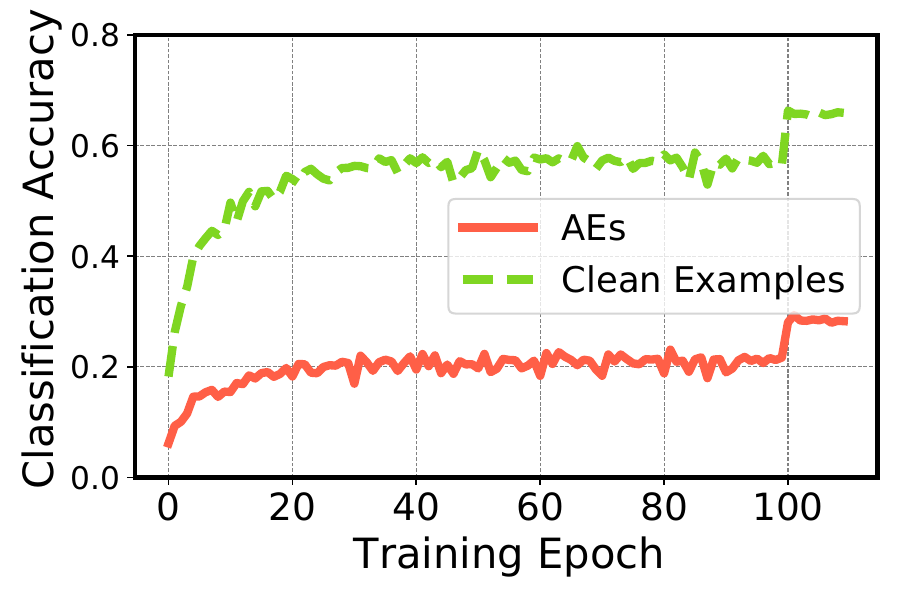}
		\caption{FGSM-PCO}
		\label{FGSM-PCO}
	\end{subfigure}
	\caption{Classification accuracy of FGSM-MEP and FGSM-PCO on WideResNet34-10 with the CIFAR100 dataset. Our method significantly prevents catastrophic overfitting.}
	\label{Wide_model}
\end{figure}

On the WideResNet34-10 model, nearly all FGSM-based methods exhibit catastrophic overfitting. Fig.~\ref{Wide_model} illustrates that FGSM-PCO effectively prevents the overfitting problem.

\section{Divergence between Adversarial and Clean Examples}
\begin{figure}[htbp]
	\centering
	\includegraphics[width=0.5\linewidth]{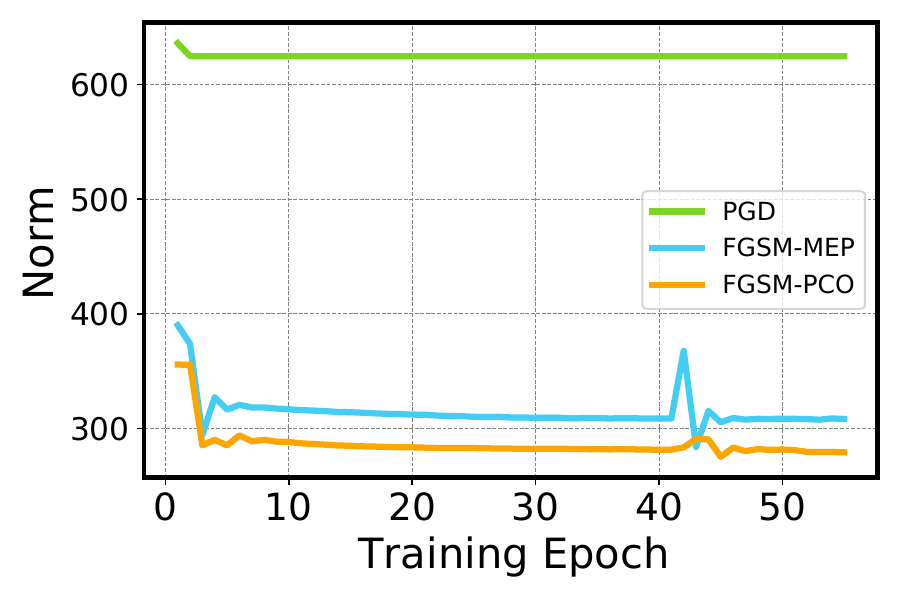 }
	\caption{The L2 norm of perturbation under different AT methods.}
	\label{norm}
\end{figure}

Apart from the classification accuracy of AT models under various attacks, the divergence between adversarial and clean examples is also a crucial metric for evaluating AT algorithms. The norm of perturbation can be regarded as a convergence criterion for the non-convex optimization problem, with smaller perturbations implying quicker convergence to local optima \cite{XiaojunJia2022PriorGuidedAI}. We evaluate the perturbation under L2 norm for PGD10, FGSM-MEP and FGSM-PCO algorithms on the CIFAR100 dataset with the ResNet18 model. The results indicate that our method achieves the smallest perturbation norm, as shown in Fig.~\ref{norm}.

\bibliographystyle{splncs04}
\bibliography{refs}